\pdfoutput=1

\documentclass[11pt]{article}
\usepackage{acl2016}
\usepackage{times}
\usepackage{caption}
\usepackage{subcaption}
\usepackage{url}
\usepackage{latexsym}
\usepackage{natbib}
\usepackage[draft]{fixme}
\usepackage{tabu}
\usepackage{booktabs}

\usepackage{graphicx} 

\usepackage{amsmath}
\usepackage{amsfonts}
\usepackage{appendix}
\usepackage{balance}
\usepackage{enumerate}

\graphicspath{{./pics/}{../pics/}}

\newtheorem{theorem}{Theorem}

\DeclareMathOperator*{\argmax}{\operatornamewithlimits{argmax}}

\aclfinalcopy 

\title{Policy Networks with Two-Stage Training for Dialogue Systems}

\author{Mehdi Fatemi \qquad Layla El Asri \qquad Hannes Schulz \qquad  Jing He \qquad  Kaheer Suleman\\ \\
 Maluuba Research \\
 Le 2000 Peel, Montr\'{e}al, QC H3A 2W5 \\
 {\tt first.last@maluuba.com}}

\date{May 11, 2016}

\begin{document}
\maketitle
\begin{abstract}
In this paper, we propose to use deep policy networks which are trained with an advantage actor-critic method for statistically optimised dialogue systems. First, we show that, on summary state and action spaces, deep Reinforcement Learning (RL) outperforms Gaussian Processes methods. Summary state and action spaces lead to good performance but require pre-engineering effort, RL knowledge, and domain expertise. In order to remove the need to define such summary spaces, we show that deep RL can also be trained efficiently on the original state and action spaces. Dialogue systems based on partially observable Markov decision processes are known to require many dialogues to train, which makes them unappealing for practical deployment. We show that a deep RL method based on an actor-critic architecture can exploit a small amount of data very efficiently. Indeed, with only a few hundred dialogues collected with a handcrafted policy, the actor-critic deep learner is considerably bootstrapped from a combination of supervised and batch RL. In addition, convergence to an optimal policy is significantly sped up compared to other deep RL methods initialized on the data with batch RL. All experiments are performed on a restaurant domain derived from the Dialogue State Tracking Challenge 2 (DSTC2) dataset.

\end{abstract}

\section{Introduction}
The statistical optimization of dialogue management in dialogue systems through Reinforcement Learning (RL) has been an active thread of research for more than two decades \citep{Levin:97,Lemon:07,Laroche:10b,Gasic:12,Daubigney:12b}. Dialogue management has been successfully modelled as a Partially Observable Markov Decision Process (POMDP) \citep{Williams:07,Gasic:12}, which leads to systems that can learn from data and which are robust to noise. In this context, a dialogue between a user and a dialogue system is framed as a sequential process where, at each turn, the system has to act based on what it has understood so far of the user's utterances.

Unfortunately, POMDP-based dialogue managers have been unfit for online deployment because they typically require several thousands of dialogues for training \citep{Gasic:10,Gasic:12}.
Nevertheless, recent work has shown that it is possible to train a POMDP-based dialogue system on just a few hundred dialogues corresponding to online interactions with users \citep{Gasic:13}. However, in order to do so, pre-engineering efforts, prior RL knowledge, and domain expertise must be applied. Indeed, summary state and action spaces must be used and the set of actions must be restricted depending on the current state so that notoriously bad actions are prohibited.

In order to alleviate the need for a summary state space, deep RL \citep{deepmind2013} has recently been applied to dialogue management \citep{Cuayahuitl:15} in the context of negotiations. It was shown that deep RL performed significantly better than other heuristic or supervised approaches. The authors performed learning over a large action space of 70 actions and they also had to use restricted action sets in order to learn efficiently over this space. Besides, deep RL was not compared to other RL methods, which we do in this paper. In \citep{cuayahuitl2016}, a simplistic implementation of deep Q Networks is presented, again with no comparison to other RL methods.

In this paper, we propose to efficiently alleviate the need for summary spaces and restricted actions using deep RL. We analyse four deep RL models: Deep Q Networks (DQN) \citep{deepmind2013}, Double DQN (DDQN) \citep{vanhasselt2015}, Deep Advantage Actor-Critic (DA2C) \citep{sutton2000} and a version of DA2C initialized with supervised learning (TDA2C)\footnote{Teacher DA2C} (similar idea to \cite{Silver:2016aa}). All models are trained on a restaurant-seeking domain. We use the Dialogue State Tracking Challenge 2 (DSTC2) dataset to train an agenda-based user simulator \citep{schatzmann2009} for online learning and to perform batch RL and supervised learning.

We first show that, on summary state and action spaces, deep RL converges faster than Gaussian Processes SARSA (GPSARSA) \citep{Gasic:10}. Then we show that deep RL enables us to work on the original state and action spaces. Although GPSARSA has also been tried on original state space \citep{Gasic:12}, it is extremely slow in terms of wall-clock time due to its growing kernel evaluations. Indeed, contrary to methods such as GPSARSA, deep RL performs efficient generalization over the state space and memory requirements do not increase with the number of experiments. On the simple domain specified by DSTC2, we do not need to restrict the actions in order to learn efficiently. In order to remove the need for restricted actions in more complex domains, we advocate for the use of TDA2C and supervised learning as a pre-training step. We show that supervised learning on a small set of dialogues (only 706 dialogues) significantly bootstraps TDA2C and enables us to start learning with a policy that already selects only valid actions, which makes for a safe user experience in deployment. Therefore, we conclude that TDA2C is very appealing for the practical deployment of POMDP-based dialogue systems.

In Section \ref{prelim} we briefly review POMDP, RL and GPSARSA. The value-based deep RL models investigated in this paper (DQN and DDQN) are described in Section \ref{sec_deepRL}. Policy networks and DA2C are discussed in Section \ref{sec_aa2c}. We then introduce the two-stage training of DA2C in Section \ref{sec_tda2c}. Experimental results are presented in Section \ref{sec_experiments}. Finally, Section \ref{sec_conclusion} concludes the paper and makes suggestions for future research.

\section{Preliminaries} \label{prelim}
The reinforcement learning problem consists of an environment (the user) and an agent (the system) \citep{sutton_RL}. The environment is described as a set of continuous or discrete states $\mathcal{S}$ and at each state $s \in \mathcal{S}$, the system can perform an action from an action space $\mathcal{A}(s)$. The actions can be continuous, but in our case they are assumed to be discrete and finite. At time $t$, as a consequence of an action $A_{t}=a\in \mathcal{A}(s)$, the state transitions from $S_{t} = s$ to $S_{t+1} = s'\in \mathcal{S}$. In addition, a reward signal $R_{t+1} = R(S_{t}, A_{t}, S_{t+1})\in \mathbb{R}$ provides feedback on the quality of the transition\footnote{In this paper, upper-case letters are used for random variables, lower-case letters for non-random values (known or unknown), and calligraphy letters for sets.}. The agent's task is to maximize at each state the expected discounted sum of rewards received after visiting this state. For this purpose, value functions are computed. The action-state value function $Q$ is defined as:
\begin{align} \label{eq_q}
\nonumber Q^{\pi}(S_{t}, A_{t}) = \mathbb{E}_{\pi}[R_{t+1} & + \gamma R_{t+2} + \gamma^2 R_{t+3} + \ldots  \\
&|~S_{t}=s, A_{t}=a],
\end{align}
where $\gamma$ is a discount factor in $[0,1]$.
In this equation, the \textit{policy} $\pi$ specifies the system's behaviour, \textit{i.e.}, it describes the agent's action selection process at each state. A policy can be a deterministic mapping $\pi(s) = a$, which specifies the action $a$ to be selected when state $s$ is met. 
On the other hand, a stochastic policy provides a probability distribution over the action space at each state:
\begin{align}
\pi(a|s) = \mathbb{P}[A_{t}=a| S_{t}=s].
\end{align}
The agent's goal is to find a policy that maximizes the $Q$-function at each state. 

It is important to note that here the system does not have direct access to the state $s$. Instead, it sees this state through a \textit{perception} process which typically includes an \textit{Automatic Speech Recognition} (ASR) step, a \textit{Natural Language Understanding} (NLU) step, and a \textit{State Tracking} (ST) step. This perception process injects noise in the state of the system and it has been shown that modelling dialogue management as a POMDP helps to overcome this noise \citep{Williams:07,young2013}.

Within the POMDP framework, the state at time $t$, $S_t,$ is not directly observable. Instead, the system has access to a noisy observation $O_t$.\footnote{Here, the representation of the user's goal and the user's utterances.} A POMDP is a tuple $(\mathcal{S}, \mathcal{A}, P, R, \mathcal{O}, Z, \gamma, b_0)$ where $\mathcal{S}$ is the state space, $\mathcal{A}$ is the action space, $P$ is the function encoding the transition probability: $P_a(s, s') = \mathbb{P}(S_{t+1}=s'~|~S_{t}=s, A_{t}=a)$, $R$ is the reward function, $\mathcal{O}$ is the observation space, $Z$ encodes the observation probabilities $Z_a(s, o) = \mathbb{P}(O_{t}=o~|~S_{t}=s, A_{t}=a)$,  $\gamma$ is a discount factor, and $b_0$ is an initial belief state. The \textit{belief state} is a distribution over states. Starting from $b_0$, the state tracker maintains and updates the belief state according to the observations perceived during the dialogue. The dialogue manager then operates on this belief state. Consequently, the value functions as well as the policy of the agent are computed on the belief states $B_t$:
 \begin{align}
Q^{\pi}(B_{t}, A_{t}) &= \mathbb{E}_{\pi}\left[\sum_{t' \geq t} \gamma^{t' - t} R_{t'+1}~|~B_t, A_t\right]  \nonumber \\
\pi(a|b) &= \mathbb{P}[A_{t}=a | B_{t}=b].
\end{align}


In this paper, we use GPSARSA as a baseline as it has been proved to be a successful algorithm for training POMDP-based dialogue managers \citep{engel2005,Gasic:10}. 
Formally, the $Q$-function is modelled as a Gaussian process, entirely defined by a mean and a kernel: $Q(B, A) \sim \mathcal{GP}(m, (k(B, A), k(B, A)))$. The mean is usually initialized at 0 and it is then jointly updated with the covariance based on the system's observations (\textit{i.e.}, the visited belief states and actions, and the rewards). In order to avoid intractability in the number of experiments, we use kernel span sparsification \citep{engel2005}. This technique consists of approximating the kernel on a dictionary of linearly independent belief states. This dictionary is incrementally built during learning. Kernel span sparsification requires setting a threshold on the precision to which the kernel is computed. As discussed in Section \ref{sec_experiments}, this threshold needs to be fine-tuned for a good tradeoff between precision and performance. 

\section{Value-Based Deep Reinforcement Learning} \label{sec_deepRL}
Broadly speaking, there are two main streams of methodologies in the RL literature: value approximation and policy gradients. As suggested by their names, the former tries to approximate the value function whereas the latter tries to directly approximate the policy. Approximations are necessary for large or continuous belief and action spaces. Indeed, if the belief space is large or continuous it would not be possible to store a value for each state in a table, so generalization over the state space is necessary. In this context, some of the benefits of deep RL techniques are the following:
\begin{itemize}
\item Generalisation over the belief space is efficient and the need for summary spaces is eliminated, normally with considerably less wall-clock training time comparing to GPSARSA, for example.
\item Memory requirements are limited and can be determined in advance unlike with methods such as GPSARSA.
\item Deep architectures with several hidden layers can be efficiently used for complex tasks and environments.
\end{itemize}

\subsection{Deep Q Networks}

A Deep $Q$-Network (DQN) is a multi-layer neural network which maps a belief state $B_t$ to the values of the possible actions $A_t \in \mathcal{A}(B_t = b)$ at that state, $Q^{\pi}(B_{t},A_{t};~ w_t)$, where $w_t$ is the weight vector of the neural network. Neural networks for the approximation of value functions have long been investigated \citep{bertsekas_neuro}. However, these methods were previously quite unstable \citep{deepmind2013}. In DQN, \citet{deepmind2013,Mnih:2015aa} proposed two techniques to overcome this instability-namely \textit{experience replay} and the use of a \textit{target network}. In experience replay, all the transitions are put in a finite pool $\mathcal{D}$ \citep{lin1993}. Once the pool has reached its predefined maximum size, adding a new transition results in deleting the oldest transition in the pool. During training, a mini-batch of transitions is \textit{uniformly} sampled from the pool, \textit{i.e.} $(B_{t}, A_{t}, R_{t+1}, B_{t+1}) \sim U(\mathcal{D})$. This method removes the instability arising from strong correlation between the subsequent transitions of an episode (a dialogue). Additionally, a target network with weight vector $w^{-}$ is used. This target network is similar to the $Q$-network except that its weights are only copied every $\tau$ steps from the $Q$-network, and remain fixed during all the other steps. The loss function for the $Q$-network at iteration $t$ takes the following form:
\begin{align} \label{eq_dqn}
\nonumber L_{t}(w_{t}) &= \mathbb{E}_{(B_{t}, A_{t}, R_{t+1}, B_{t+1})\sim U(\mathcal{D})} \Big[ \\
\nonumber \Big(&R_{t+1} + \gamma \max_{a'} Q^{\pi}(B_{t+1}, a'; w^{-}_{t})  \\
&- Q^{\pi}(B_{t}, A_{t}; w_{t}) \Big)^{2} ~\Big].
\end{align}

\subsection{Double DQN: Overcoming Overestimation and  Instability of DQN}

The \emph{max} operator in Equation \ref{eq_dqn} uses the same value network (\textit{i.e.}, the target network) to select actions and evaluate them. This increases the probability of overestimating the value of the state-action pairs \citep{vanhasselt2010,vanhasselt2015}. To see this more clearly, the target part of the loss in Equation \ref{eq_dqn} can be rewritten as follows:
\begin{align} 
\nonumber R_{t+1} + \gamma Q^{\pi}(B_{t+1}, \argmax_{a}{Q^{\pi}(B_{t+1}, a; w^{-}_{t})}; w^{-}_{t}).
\end{align}
In this equation, the target network is used twice. Decoupling is possible by using the $Q$-network for action selection as follows \citep{vanhasselt2015}:
\begin{align} 
\nonumber R_{t+1} + \gamma Q^{\pi}(B_{t+1}, \argmax_{a}{Q^{\pi}(B_{t+1}, a; w_{t})}; w^{-}_{t}).
\end{align}
Then, similarly to DQN, the $Q$-network is trained using experience replay and the target network is updated every $\tau$ steps. This new version of DQN, called Double DQN (DDQN), uses the two value networks in a decoupled manner, and alleviates the overestimation issue of DQN. This generally results in a more stable learning process \citep{vanhasselt2015}.

In the following section, we present deep RL models which perform policy search and output a stochastic policy rather than value approximation with a deterministic policy.
\section{Policy Networks and Deep Advantage Actor-Critic (DA2C)} \label{sec_aa2c}

A policy network is a parametrized probabilistic mapping between belief and action spaces:
\begin{align}
\nonumber\pi_{\theta}(a|b) = \pi(a|b; \theta) = \mathbb{P}(A_{t} = a | B_{t}=b, \theta_{t}=\theta),
\end{align}
where $\theta$ is the parameter vector (the weight vector of a neural network).\footnote{For parametrization, we use $w$ for value networks and $\theta$ for policy networks.} In order to train policy networks, policy gradient algorithms have been developed \citep{williams1992,sutton2000}. Policy gradient algorithms are model-free methods which directly approximate the policy by parametrizing it. The parameters are learnt using a gradient-based optimization method. 

We first need to define an objective function $J$ that will lead the search for the parameters $\theta$. This objective function defines policy quality. One way of defining it is to take the average over the rewards received by the agent. Another way is to compute the discounted sum of rewards for each trajectory, given that there is a designated start state. The policy gradient is then computed according to the \textit{Policy Gradient Theorem} \citep{sutton2000}.
\begin{theorem}[Policy Gradient]
For any differentiable policy $\pi_{\theta}(b, a)$ and for the average reward or the start-state objective function, the policy gradient can be computed as
\begin{align} \label{eq_policy_grad}
\nabla_{\theta}J(\theta) = \mathbb{E}_{\pi_{\theta}}[ \nabla_{\theta} \log \pi_{\theta}(a|b) Q^{\pi_{\theta}}(b, a) ].
\end{align}
\end{theorem}
Policy gradient methods have been used successfully in different domains. Two recent examples are AlphaGo by DeepMind \citep{Silver:2016aa} and MazeBase by Facebook AI \citep{sukhbaatar2016}.

One way to exploit Theorem 1 is to parametrize $Q^{\pi_{\theta}}(b, a)$ separately (with a parameter vector $w$) and learn the parameter vector during training in a similar way as in DQN. The trained $Q$-network can then be used for policy evaluation in Equation \ref{eq_policy_grad}. Such algorithms are known in general as \emph{actor-critic} algorithms, where the $Q$ approximator is the critic and $\pi_{\theta}$ is the actor \citep{sutton_phd,Barto:90,Bhatnagar:09}. This can be achieved with \textit{two} separate deep neural networks: a \emph{$Q$-Network} and a \emph{policy network}.

However, a direct use of Equation \ref{eq_policy_grad} with $Q$ as critic is known to cause high variance \citep{williams1992}. An important property of Equation \ref{eq_policy_grad} can be used in order to overcome this issue: subtracting any differentiable function $Ba$ expressed over the belief space from $Q^{\pi_{\theta}}$ will not change the gradient.
A good selection of $Ba$, which is called the \emph{baseline}, can reduce the variance dramatically \citep{sutton_RL}. As a result, Equation \ref{eq_policy_grad} may be rewritten as follows:
\begin{align} \label{eq_advantage}
\nabla_{\theta}J(\theta) = \mathbb{E}_{\pi_{\theta}}[ \nabla_{\theta} \log \pi_{\theta}(a|b) Ad(b, a)],
\end{align}
where $Ad(b,a) = Q^{\pi_{\theta}}(b,a) - Ba(b)$ is called the \emph{advantage function}. A good baseline is the value function $V^{\pi_{\theta}}$, for which the advantage function becomes $Ad(b,a) = Q^{\pi_{\theta}}(b,a) - V^{\pi_{\theta}}(b)$. However, in this setting, we need to train two separate networks to parametrize $Q^{\pi_{\theta}}$ and $V^{\pi_{\theta}}$. A better approach is to use the TD error $\delta = R_{t+1} + \gamma V^{\pi_{\theta}}(B_{t+1}) - V^{\pi_{\theta}}(B_{t})$ as advantage function. It can be proved that the expected value of the TD error is $Q^{\pi_{\theta}}(b,a) - V^{\pi_{\theta}}(b)$. If the TD error is used, only one network is needed, to parametrize $V^{\pi_{\theta}}(B_{t}) = V^{\pi_{\theta}}(B_{t}; w_{t})$. We call this network the \emph{value network}. We can use a DQN-like method to train the value network using both experience replay and a target network. For a transition $B_{t}=b$, $A_{t}=a$, $R_{t+1}=r$ and $B_{t+1}=b'$, the advantage function is calculated as in:
\begin{align}
\delta_{t} = r + \gamma V^{\pi_{\theta}}(b'; w_{t}) - V^{\pi_{\theta}}(b; w_{t}).
\end{align}
Because the gradient in Equation \ref{eq_advantage} is weighted by the advantage function, it may become quite large. In fact, the advantage function may act as a large learning rate. This can cause the learning process to become unstable. To avoid this issue, we add $L_{2}$ regularization to the policy objective function. We call this method Deep Advantage Actor-Critic (DA2C). 


\begin{figure*}[!t]
\begin{center}
\begin{subfigure}[t]{3in}
	\includegraphics[trim = 0in 0 0 0in, clip, width=3in, height = 3in]{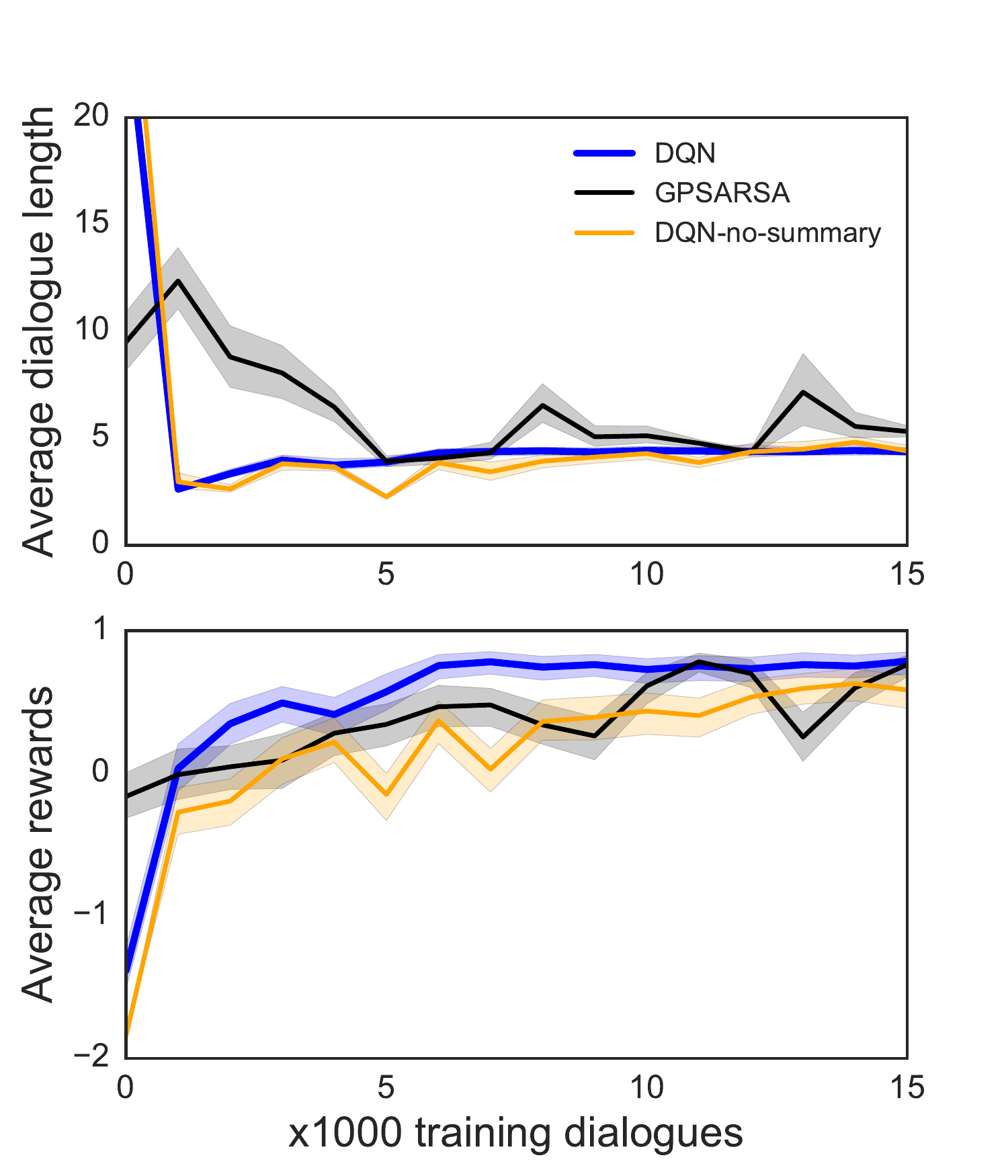}
	\caption{Comparison of GPSARSA  on summary spaces and DQN on summary (DQN) and original spaces (DQN-no-summary).}
	\label{dqn_gp}
\end{subfigure}
\quad
\begin{subfigure}[t]{3in}
	\includegraphics[trim = 0in 0 0 0in, clip, width=3in, height = 3in]{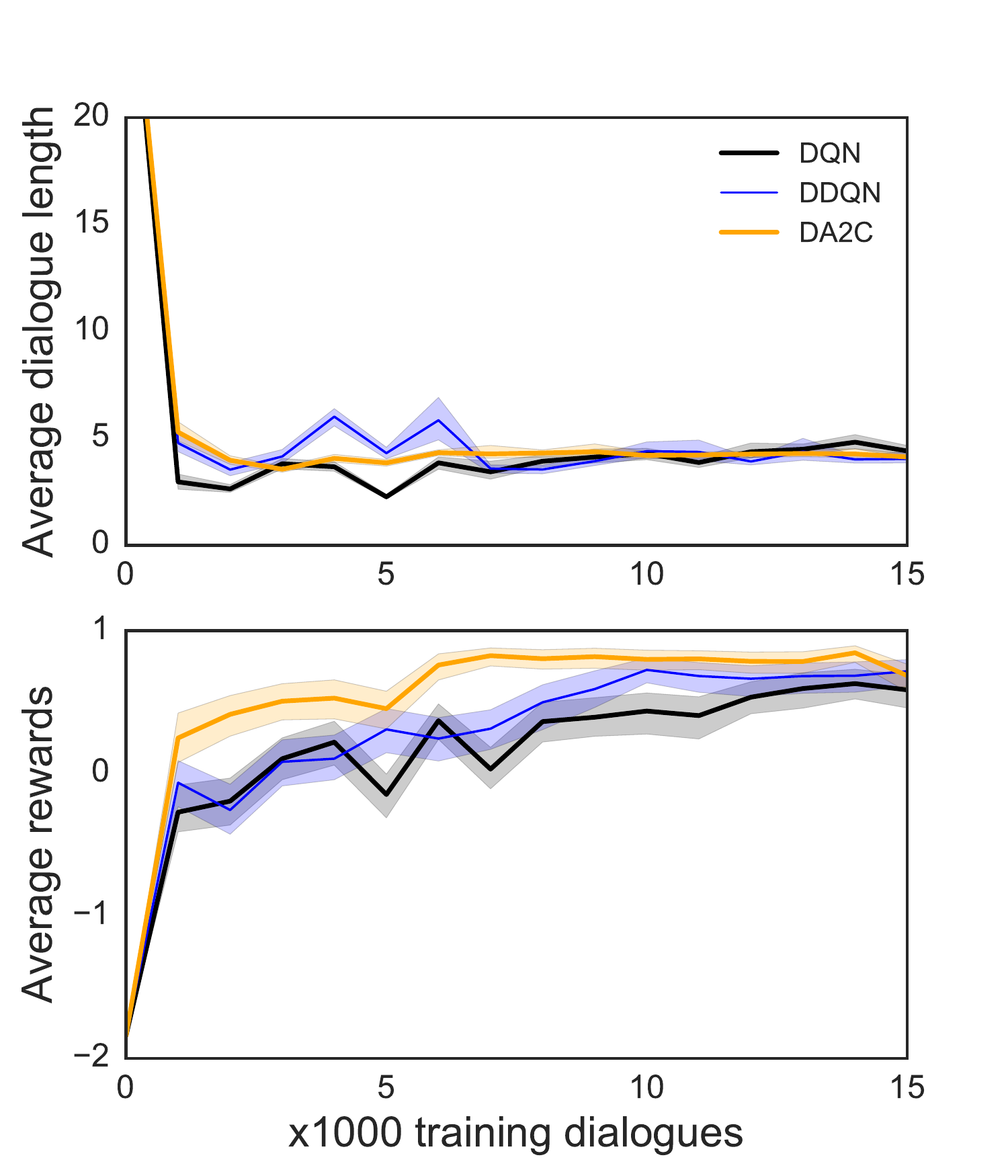}
	\caption{Comparison of DA2C, DQN and DDQN on original spaces.}
	\label{no_batch}
\end{subfigure}
\caption{Comparison of different algorithms on simulated dialogues, without any pre-training.}
\label{fig_dqn_gp}
\end{center}
\end{figure*}

In the next section, we show how this architecture can be used to efficiently exploit a small set of handcrafted data.
\section{Two-stage Training of the Policy Network} \label{sec_tda2c}
By definition, the policy network provides a probability distribution over the action space. As a result and in contrast to value-based methods such as DQN, a policy network can also be trained with direct \emph{supervised learning} \citep{Silver:2016aa}.
Supervised training of RL agents has been well-studied in the context of Imitation Learning (IL). In IL, an agent learns to reproduce the behaviour of an expert. Supervised learning of the policy was one of the first techniques used to solve this problem \citep{Pomerleau:89,Amit:02}. This direct type of imitation learning requires that the learning agent and the expert share the same characteristics. If this condition is not met, IL can be done at the level of the value functions rather than the policy directly \citep{Piot:15}. In this paper, the data that we use (DSTC2) was collected with a dialogue system similar to the one we train so in our case, the demonstrator and the learner share the same characteristics. 

Similarly to \citet{Silver:2016aa}, here, we initialize both the policy network and the value network on the data. The policy network is trained by minimising the categorical cross-entropy between the predicted action distribution and the demonstrated actions. The value network is trained directly through RL rather than IL to give more flexibility in the kind of data we can use. Indeed, our goal is to collect a small number of dialogues and learn from them. IL usually assumes that the data corresponds to expert policies. However, dialogues collected with a handcrafted policy or in a Wizard-of-Oz (WoZ) setting often contain both optimal and sub-optimal dialogues and RL can be used to learn from all of these dialogues. Supervised training can also be done on these dialogues as we show in Section \ref{sec_experiments}.

Supervised actor-critic architectures following this idea have been proposed in the past \citep{Benbrahim:97,Si:04}; the actor works together with a human supervisor to gain competence on its task even if the critic's estimations are poor. For instance, a human can help a robot move by providing the robot with valid actions. We advocate for the same kind of methods for dialogue systems. It is easy to collect a small number of high-quality dialogues and then use supervised learning on this data to teach the system valid actions. This also eliminates the need to define restricted action sets.

In all the methods above, \emph{Adadelta} will be used as the gradient-decent optimiser, which in our experiments works noticeably better than other methods such as \emph{Adagrad}, \emph{Adam}, and \emph{RMSProp}. 

\section{Experiments}  \label{sec_experiments}

\subsection{Comparison of DQN and GPSARSA}

\subsubsection{Experimental Protocol}
In this section, as a first argument in favour of deep RL, we perform a comparison between GPSARSA and DQN on simulated dialogues. We trained an agenda-based user simulator which at each dialogue turn, provides one or several dialogue act(s) in response to the latest machine act \citep{schatzmann2007,schatzmann2009}. The dataset used for training this user-simulator is the Dialogue State Tracking Challenge 2 (DSTC2) \citep{Henderson14a} dataset. State tracking is also trained on this dataset. DSTC2 includes dialogues with users who are searching for restaurants in Cambridge, UK.

In each dialogue, the user has a goal containing constraint slots and request slots. The constraint and request slots available in DSTC2 are listed in Appendix A. The constraints are the slots that the user has to provide to the system (for instance the user is looking for a specific type of food in a given area) and the requests are the slots that the user must receive from the system (for instance the user wants to know the address and phone number of the restaurant found by the system).

Similarly, the belief state is composed of two parts: constraints and requests. The constraint part includes the probabilities of the top two values for each constraint slot as returned by the state tracker (the value might be empty with a probability zero if the slot has not been mentioned). The request part, on the other hand, includes the probability of each request slot. For instance the constraint part might be [food: (Italian, 0.85) (Indian, 0.1) (Not\_mentioned, 0.05)] and the request part might be [area: 0.95] meaning that the user is probably looking for an Italian restaurant and that he wants to know the area of the restaurant found by the system. To compare DQN to GPSARSA, we work on a summary state space \citep{Gasic:12,Gasic:13}. Each constraint is mapped to a one-hot vector, with 1 corresponding to the tuple in the grid vector $g_{c} = [(1, 0), (.8, .2), (.6, .2), (.6, .4), (.4, .4)]$ that minimizes the Euclidean distance to the top two probabilities. Similarly, each request slot is mapped to a one-hot vector according to the grid $g_{r} = [1, .8, .6, .4, 0.]$. The final belief vector, known as the summary state, is defined as the concatenation of the constraint and request one-hot vectors. Each summary state is a binary vector of length 60 (12 one-hot vectors of length 5) and the total number of states is $5^{12}$.

We also work on a summary action space and we use the act types listed in Table~\ref{tab:summary_actions} in Appendix A. We add the necessary slot information as a post processing step. For example, the \textit{request} act means that the system wants to request a slot from the user, \textit{e.g.} request(food). In this case, the selection of the slot is based on min-max probability, \textit{i.e.}, the most ambiguous slot (which is the slot we want to request) is assumed to be the one for which the value with maximum probability has the minimum probability compared to the most certain values of the other slots. Note that this heuristic approach to compute the summary state and action spaces is a requirement to make GPSARSA tractable; it is a serious limitation in general and should be avoided.

As reward, we use a normalized scheme with a reward of +1 if the dialogue finishes successfully before 30 turns,\footnote{A dialogue is successful if the user retrieves all the request slots for a restaurant matching all the constraints of his goal.} a reward of -1 if the dialogue is not successful after 30 turns, and a reward of -0.03 for each turn. A reward of -1 is also distributed to the system if the user hangs up. In our settings, the user simulator hangs up every time the system proposes a restaurant which does not match at least one of his constraints.

For the deep $Q$-network, a Multi-Layer Perceptron (MLP) is used with two fully connected hidden layers, each having a \textit{tanh} activation. The output layer has no activation and it provides the value for each of the summary machine acts. The summary machine acts are mapped to original acts using the heuristics explained previously. Both algorithms are trained with 15000 dialogues. GPSARSA is trained with $\epsilon$-\textit{softmax} exploration, which, with probability $1-\epsilon$, selects an action based on the logistic distribution $\mathbb{P}[a| b] = \frac{e^{Q(b, a)}}{\sum_{a'}{e^{Q(b, a')}}}$ and, with probability $\epsilon$, selects an action in a uniformly random way. From our experiments, this exploration scheme works best in terms of both convergence rate and variance. For DQN, we use a simple $\epsilon$-\textit{greedy} exploration which, with probability $\epsilon$ (same $\epsilon$ as above), uniformly selects an action and, with probability $1 - \epsilon$, selects an action maximizing the $Q$-function. For both algorithms, $\epsilon$ is annealed to less than 0.1 over the course of training.

In a second experiment, we remove both summary state and action spaces for DQN, \textit{i.e.}, we do not perform the Euclidean-distance mapping as before but instead work directly on the probabilities themselves. Additionally, the state is augmented with the probability (returned by the state tracker) of each user act (see Table~\ref{tab:user_acts} in Appendix A), the dialogue turn, and the number of results returned by the database (0 if there was no query). Consequently, the state consists of 31 continuous values and two discrete values. The original action space is composed of 11 actions: \texttt{offer}\footnote{This act consists of proposing a restaurant to the user. In order to be consistent with the DSTC2 dataset, an \texttt{offer} always contains the values for all the constraints understood by the system, \textit{e.g.} offer(name = Super Ramen, food = Japanese, price range = cheap).}, \texttt{select-area}, \texttt{select-food}, \texttt{select-pricerange}, \texttt{request-area}, \texttt{request-food}, \texttt{request-pricerange}, \texttt{expl-conf-area}, \texttt{expl-conf-food}, \texttt{expl-conf-pricerange}, \texttt{repeat}. There is no post-processing via min-max selection anymore since the slot is part of the action, \textit{e.g.}, \texttt{select-area}.

The policies are evaluated after each 1000 training dialogues on 500 test dialogues without exploration. 

\subsubsection{Results}
Figure \ref{fig_dqn_gp} illustrates the performance of DQN compared to GPSARSA. In our experiments with GPSARSA we found that it was difficult to find a good tradeoff between precision and efficiency. Indeed, for low precision, the algorithm learned rapidly but did not reach optimal behaviour, whereas higher precision made learning extremely slow but resulted in better end-performance. On summary spaces, DQN outperforms GPSARSA in terms of convergence. Indeed, GPSARSA requires twice as many dialogues to converge. It is also worth mentioning here that the wall-clock training time of GPSARSA is considerably longer than the one of DQN due to kernel evaluation. The second experiment validates the fact that Deep RL can be efficiently trained directly on the belief state returned by the state tracker. Indeed, DQN on the original spaces performs as well as GPSARSA on the summary spaces. 

In the next section, we train and compare the deep RL networks previously described on the original state and action spaces.

\subsection{Comparison of the Deep RL Methods}

\begin{figure*}[!t]
\begin{center}
\begin{subfigure}[t]{3in}
	\includegraphics[trim = 0in 0 0 0in, clip, width=3in, height = 3in]{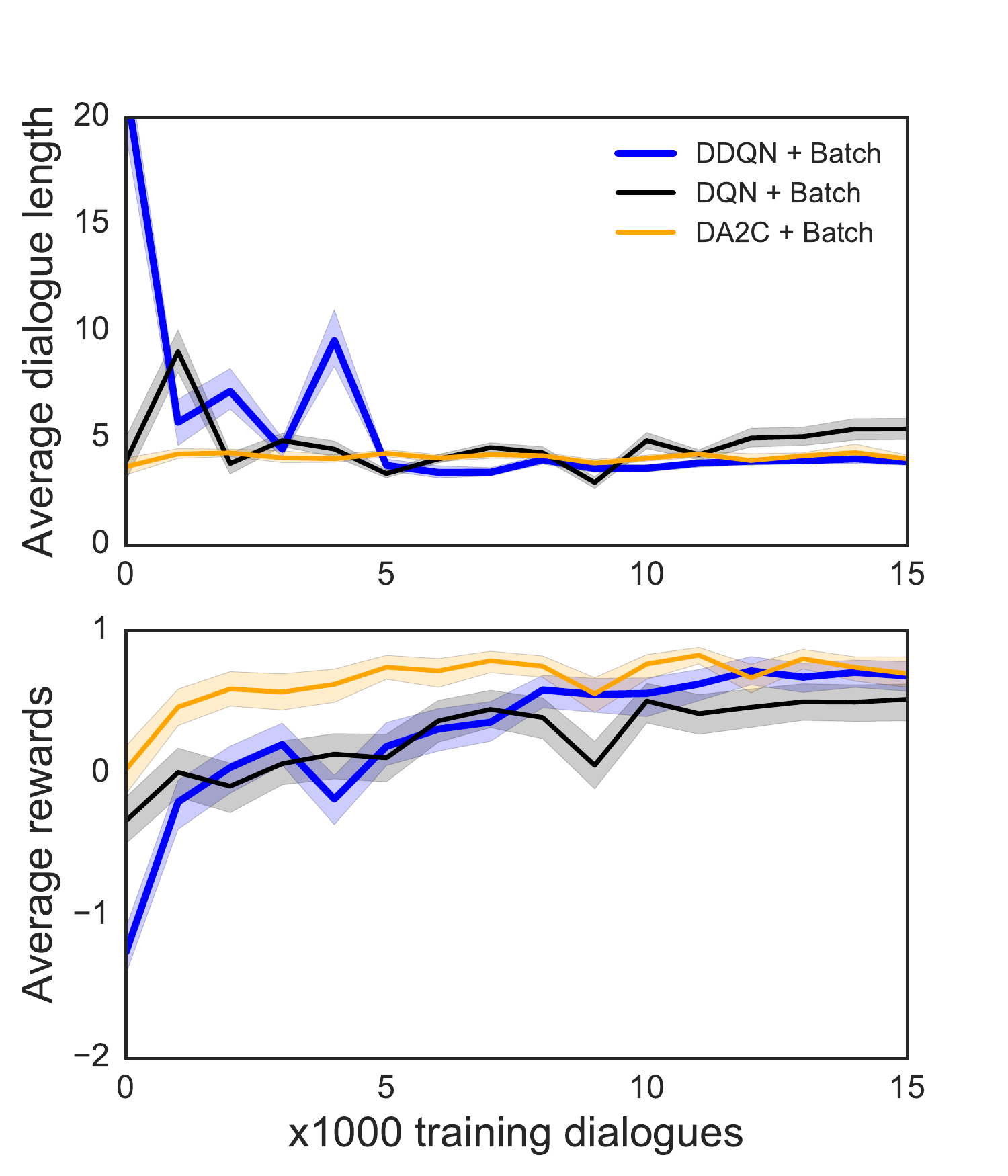}
	\caption{Comparison of DA2C, DQN and DDQN after batch initialization.}
	\label{batch}
\end{subfigure}
\quad
\begin{subfigure}[t]{3in}
	\includegraphics[trim = 0in 0 0 0in, clip, width=3in, height = 3in]{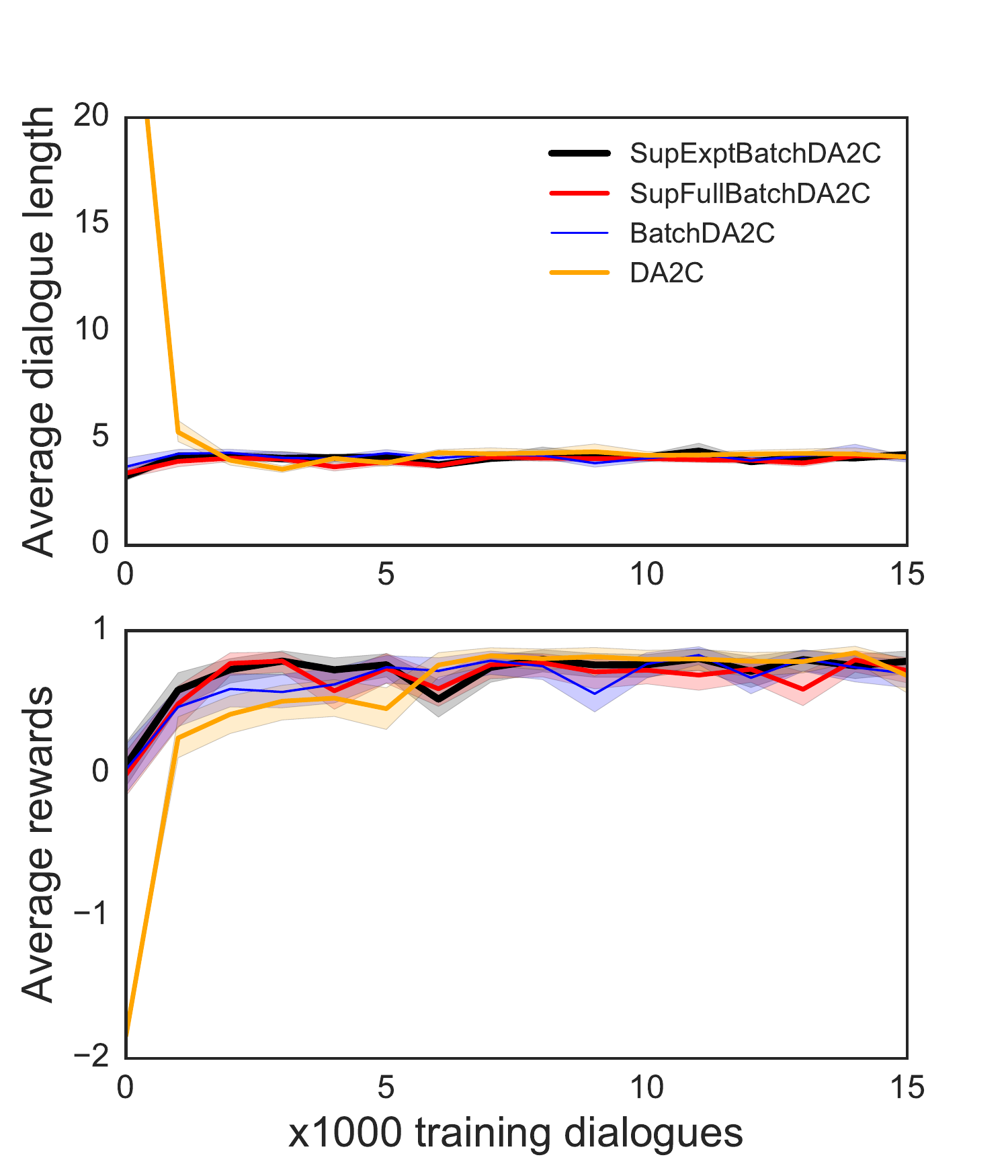}
	\caption{Comparison of DA2C and DA2C after batch initialization (batchDA2C), and TDA2C after supervised training on expert (SupExptBatchDA2C) and non-expert data (SupFullBatchDA2C).}
	\label{da2c}
\end{subfigure}
\caption{Comparison of different algorithms on simulated dialogues, with pre-training.}
\label{fig_comparison}
\end{center}
\end{figure*}

\subsubsection{Experimental Protocol}
Similarly to the previous example, we work on a restaurant domain and use the DSTC2 specifications. We use $\epsilon-$greedy exploration for all four algorithms with $\epsilon$ starting at 0.5 and being linearly annealed at a rate of $\lambda=0.99995$. To speed up the learning process, the actions \texttt{select-pricerange}, \texttt{select-area}, and \texttt{select-food} are excluded from exploration. Note that this set does not depend on the state and is meant for exploration only. All the actions can be performed by the system at any moment.

We derived two datasets from DSTC2. The first dataset contains the 2118 dialogues of DSTC2. We had these dialogues rated by a human expert, based on the quality of dialogue management and on a scale of 0 to 3. The second dataset only contains the dialogues with a rating of 3 (706 dialogues). The underlying assumption is that these dialogues correspond to optimal policies. 

We compare the convergence rates of the deep RL models in different settings. First, we compare DQN, DDQN and DA2C without any pre-training (Figure \ref{no_batch}). Then, we compare DQN, DDQN and TDA2C with an RL initialization on the DSTC2 dataset (Figure \ref{batch}). Finally, we focus on the advantage actor-critic models and compare DA2C, TDA2C, TDA2C with batch initialization on DSTC2, and TDA2C with batch initialization on the expert dialogues (Figure \ref{da2c}).

\subsubsection{Results}
As expected, DDQN converges faster than DQN on all experiments. Figure \ref{no_batch} shows that, without any pre-training, DA2C is the one which converges the fastest (6000 dialogues \textit{vs.} 10000 dialogues for the other models). Figure \ref{batch} gives consistent results and shows that, with initial training on the 2118 dialogues of DSTC2, TDA2C converges significantly faster than the other models. Figure \ref{da2c} focuses on DA2C and TDA2C. Compared to batch training, supervised training on DSTC2 speeds up convergence by 2000 dialogues (3000 dialogues \textit{vs.} 5000 dialogues). Interestingly, there does not seem to be much difference between supervised training on the expert data and on DSTC2. The expert data only consists of 706 dialogues out of 2118 dialogues. Our observation is that, in the non-expert data, many of the dialogue acts chosen by the system were still appropriate, which explains that the system learns acceptable behavior from the entire dataset. This shows that supervised training, even when performed not only on optimal dialogues, makes learning much faster and relieves the need for restricted action sets. Valid actions are learnt from the dialogues and then RL exploits the good and bad dialogues to pursue training towards a high performing policy.


\section{Concluding Remarks} \label{sec_conclusion}
In this paper, we used policy networks for dialogue systems and trained them in a two-stage fashion: supervised training and batch reinforcement learning followed by online reinforcement learning. An important feature of policy networks is that they directly provide a probability distribution over the action space, which enables supervised training. We compared the results with other deep reinforcement learning algorithms, namely Deep Q Networks and Double Deep Q Networks. The combination of supervised and reinforcement learning is the main benefit of our method, which paves the way for developing trainable end-to-end dialogue systems. Supervised training on a small dataset considerably bootstraps the learning process and can be used to significantly improve the convergence rate of reinforcement learning in statistically optimised dialogue systems.

\bibliographystyle{acl_natbib}
\bibliography{bibfile}

\begin{thebibliography}{}
\expandafter\ifx\csname natexlab\endcsname\relax\def\natexlab#1{#1}\fi

\bibitem[{Amit and Mataric(2002)}]{Amit:02}
R.~Amit and M.~Mataric. 2002.
\newblock Learning movement sequences from demonstration.
\newblock In {\em Proc. Int. Conf. on Development and Learning\/}. pages
  203--208.

\bibitem[{Barto et~al.(1990)Barto, Sutton, and Anderson}]{Barto:90}
A.~G. Barto, R.~S. Sutton, and C.~W. Anderson. 1990.
\newblock In {\em Artificial Neural Networks\/}, chapter Neuronlike Adaptive
  Elements That Can Solve Difficult Learning Control Problems, pages 81--93.

\bibitem[{Benbrahim and Franklin(1997)}]{Benbrahim:97}
H.~Benbrahim and J.~A. Franklin. 1997.
\newblock Biped dynamic walking using reinforcement learning.
\newblock {\em Robotics and Autonomous Systems\/} 22:283--302.

\bibitem[{Bertsekas and Tsitsiklis(1996)}]{bertsekas_neuro}
D.~P. Bertsekas and J.~Tsitsiklis. 1996.
\newblock {\em Neuro-Dynamic Programming\/}.
\newblock Athena Scientific.

\bibitem[{Bhatnagar et~al.(2009)Bhatnagar, Sutton, Ghavamzadeh, and
  Lee}]{Bhatnagar:09}
S.~Bhatnagar, R.~Sutton, M.~Ghavamzadeh, and M.~Lee. 2009.
\newblock {Natural Actor-Critic Algorithms}.
\newblock {\em {Automatica}\/} 45(11).

\bibitem[{Cuay{\'{a}}huitl(2016)}]{cuayahuitl2016}
H.~Cuay{\'{a}}huitl. 2016.
\newblock Simpleds: A simple deep reinforcement learning dialogue system.
\newblock {arXiv:1601.04574v1 [cs.AI]}.

\bibitem[{Cuay{\'{a}}huitl et~al.(2015)Cuay{\'{a}}huitl, Keizer, and
  Lemon}]{Cuayahuitl:15}
H.~Cuay{\'{a}}huitl, S.~Keizer, and O.~Lemon. 2015.
\newblock Strategic dialogue management via deep reinforcement learning.
\newblock {arXiv:1511.08099 [cs.AI]}.

\bibitem[{Daubigney et~al.(2012)Daubigney, Geist, Chandramohan, and
  Pietquin}]{Daubigney:12b}
L.~Daubigney, M.~Geist, S.~Chandramohan, and O.~Pietquin. 2012.
\newblock {A Comprehensive Reinforcement Learning Framework for Dialogue
  Management Optimisation}.
\newblock {\em IEEE Journal of Selected Topics in Signal Processing\/}
  6(8):891--902.

\bibitem[{Engel et~al.(2005)Engel, Mannor, and Meir}]{engel2005}
Y.~Engel, S.~Mannor, and R.~Meir. 2005.
\newblock Reinforcement learning with gaussian processes.
\newblock In {\em Proc. of ICML\/}.

\bibitem[{Ga\v{s}i\'{c} et~al.(2013)Ga\v{s}i\'{c}, Breslin, Henderson, Kim,
  Szummer, Thomson, Tsiakoulis, and Young}]{Gasic:13}
M.~Ga\v{s}i\'{c}, C.~Breslin, M.~Henderson, D.~Kim, M.~Szummer, B.~Thomson,
  P.~Tsiakoulis, and S.J. Young. 2013.
\newblock On-line policy optimisation of bayesian spoken dialogue systems via
  human interaction.
\newblock In {\em Proc. of ICASSP\/}. pages 8367--8371.

\bibitem[{Ga\v{s}i\'{c} et~al.(2012)Ga\v{s}i\'{c}, Henderson, Thomson,
  Tsiakoulis, and Young}]{Gasic:12}
M.~Ga\v{s}i\'{c}, M.~Henderson, B.~Thomson, P.~Tsiakoulis, and S.~Young. 2012.
\newblock Policy optimisation of {POMDP}-based dialogue systems without state
  space compression.
\newblock In {\em Proc. of SLT\/}.

\bibitem[{Ga\v{s}i\'{c} et~al.(2010)Ga\v{s}i\'{c}, Jur\v{c}\'{\i}\v{c}ek,
  Keizer, Mairesse, Thomson, Yu, and Young}]{Gasic:10}
M.~Ga\v{s}i\'{c}, F.~Jur\v{c}\'{\i}\v{c}ek, S.~Keizer, F.~Mairesse, B.~Thomson,
  K.~Yu, and S.~Young. 2010.
\newblock Gaussian processes for fast policy optimisation of {POMDP}-based
  dialogue managers.
\newblock In {\em Proc. of SIGDIAL\/}.

\bibitem[{Henderson et~al.(2014)Henderson, Thomson, and
  Williams}]{Henderson14a}
M.~Henderson, B.~Thomson, and J.~Williams. 2014.
\newblock {The Second Dialog State Tracking Challenge}.
\newblock In {\em Proc. of SIGDIAL\/}.

\bibitem[{Laroche et~al.(2010)Laroche, Putois, and Bretier}]{Laroche:10b}
R.~Laroche, G.~Putois, and P.~Bretier. 2010.
\newblock Optimising a handcrafted dialogue system design.
\newblock In {\em Proc. of Interspeech\/}.

\bibitem[{Lemon and Pietquin(2007)}]{Lemon:07}
O.~Lemon and O.~Pietquin. 2007.
\newblock Machine learning for spoken dialogue systems.
\newblock In {\em Proc. of Interspeech\/}. pages 2685--2688.

\bibitem[{Levin et~al.(1997)Levin, Pieraccini, and Eckert}]{Levin:97}
E.~Levin, R.~Pieraccini, and W.~Eckert. 1997.
\newblock Learning dialogue strategies within the markov decision process
  framework.
\newblock In {\em Proc. of ASRU\/}.

\bibitem[{Lin(1993)}]{lin1993}
L-J Lin. 1993.
\newblock {\em Reinforcement learning for robots using neural networks\/}.
\newblock Ph.D. thesis, Carnegie Mellon University.

\bibitem[{Mnih et~al.(2013)Mnih, Kavukcuoglu, Silver, Graves, Antonoglou,
  Wierstra, and Riedmiller}]{deepmind2013}
V~Mnih, K.~Kavukcuoglu, D.~Silver, A.~Graves, I~Antonoglou, D.~Wierstra, and
  M.~Riedmiller. 2013.
\newblock Playing {Atari} with deep reinforcement learning.
\newblock In {\em NIPS Deep Learning Workshop\/}.

\bibitem[{Mnih et~al.(2015)Mnih, Kavukcuoglu, Silver, Rusu, Veness, Bellemare,
  Graves, Riedmiller, Fidjeland, Ostrovski, Petersen, Beattie, Sadik,
  Antonoglou, King, Kumaran, Wierstra, Legg, and Hassabis}]{Mnih:2015aa}
V.~Mnih, K.~Kavukcuoglu, D.~Silver, A.A. Rusu, J.~Veness, M.G. Bellemare,
  A.~Graves, M.~Riedmiller, A.K. Fidjeland, G.~Ostrovski, S.~Petersen,
  C.~Beattie, A.~Sadik, I.~Antonoglou, H.~King, D.~Kumaran, D.~Wierstra,
  S.~Legg, and D.~Hassabis. 2015.
\newblock Human-level control through deep reinforcement learning.
\newblock {\em Nature\/} 518(7540):529--533.

\bibitem[{Piot et~al.(2015)Piot, Geist, and Pietquin}]{Piot:15}
B.~Piot, M.~Geist, and O.~Pietquin. 2015.
\newblock {Imitation Learning Applied to Embodied Conversational Agents}.
\newblock In {\em Proc. of MLIS\/}.

\bibitem[{Pomerleau(1989)}]{Pomerleau:89}
D.~A. Pomerleau. 1989.
\newblock Alvinn: An autonomous land vehicle in a neural network.
\newblock In {\em Proc. of NIPS\/}. pages 305--313.

\bibitem[{Schatzmann et~al.(2007)Schatzmann, Thomson, Weilhammer, Ye, and
  Young}]{schatzmann2007}
J.~Schatzmann, B.~Thomson, K.~Weilhammer, H.~Ye, and S.~Young. 2007.
\newblock Agenda-based user simulation for bootstrapping a {POMDP} dialogue
  system.
\newblock In {\em Proc. of NAACL HLT\/}. pages 149--152.

\bibitem[{Schatzmann and Young(2009)}]{schatzmann2009}
J.~Schatzmann and S.~Young. 2009.
\newblock The hidden agenda user simulation model.
\newblock {\em Proc. of TASLP\/} 17(4):733--747.

\bibitem[{Si et~al.(2004)Si, Barto, Powell, and Wunsch}]{Si:04}
J.~Si, A.~G. Barto, W.~B. Powell, and D.~Wunsch. 2004.
\newblock {\em Supervised ActorCritic Reinforcement Learning\/}, pages
  359--380.

\bibitem[{Silver et~al.(2016)Silver, Huang, Maddison, Guez, Sifre, van~den
  Driessche, Schrittwieser, Antonoglou, Panneershelvam, Lanctot, Dieleman,
  Grewe, Nham, Kalchbrenner, Sutskever, Lillicrap, Leach, Kavukcuoglu, Graepel,
  and Hassabis}]{Silver:2016aa}
D.~Silver, A.~Huang, C.J. Maddison, A.~Guez, L.~Sifre, G.~van~den Driessche,
  J.~Schrittwieser, I.~Antonoglou, V.~Panneershelvam, M.~Lanctot, S.~Dieleman,
  D.~Grewe, J.~Nham, N.~Kalchbrenner, I.~Sutskever, T.~Lillicrap, M.~Leach,
  K.~Kavukcuoglu, T.~Graepel, and D.~Hassabis. 2016.
\newblock Mastering the game of go with deep neural networks and tree search.
\newblock {\em Nature\/} 529(7587):484--489.

\bibitem[{Sukhbaatar et~al.(2016)Sukhbaatar, Szlam, Synnaeve, Chintala, and
  Fergus}]{sukhbaatar2016}
S.~Sukhbaatar, A.~Szlam, G.~Synnaeve, S.~Chintala, and R.~Fergus. 2016.
\newblock Mazebase: A sandbox for learning from games.
\newblock {arxiv.org/pdf/1511.07401 [cs.LG]}.

\bibitem[{Sutton(1984)}]{sutton_phd}
R.~S. Sutton. 1984.
\newblock {\em Temporal credit assignment in reinforcement learning\/}.
\newblock Ph.D. thesis, University of Massachusetts at Amherst, Amherst, MA,
  USA.

\bibitem[{Sutton et~al.(2000)Sutton, McAllester, Singh, and
  Mansour}]{sutton2000}
R.~S. Sutton, D.~McAllester, S.~Singh, and Y.~Mansour. 2000.
\newblock Policy gradient methods for reinforcement learning with function
  approximation.
\newblock In {\em Proc. of NIPS\/}. volume~12, pages 1057--1063.

\bibitem[{Sutton and Barto(1998)}]{sutton_RL}
R.S. Sutton and A.G. Barto. 1998.
\newblock {\em Reinforcement Learning\/}.
\newblock MIT Press.

\bibitem[{{van Hasselt}(2010)}]{vanhasselt2010}
H.~{van Hasselt}. 2010.
\newblock Double q-learning.
\newblock In {\em Proc. of NIPS\/}. pages 2613--2621.

\bibitem[{{van Hasselt} et~al.(2015){van Hasselt}, Guez, and
  Silver}]{vanhasselt2015}
H.~{van Hasselt}, A.~Guez, and D.~Silver. 2015.
\newblock Deep reinforcement learning with double {Q}-learning.
\newblock {arXiv:1509.06461v3 [cs.LG]}.

\bibitem[{Williams and Young(2007)}]{Williams:07}
J.D. Williams and S.~Young. 2007.
\newblock Partially observable markov decision processes for spoken dialog
  systems.
\newblock {\em Proc. of CSL\/} 21:231--422.

\bibitem[{Williams(1992)}]{williams1992}
R.J. Williams. 1992.
\newblock Simple statistical gradient-following algorithms for connectionist
  reinforcement learning.
\newblock {\em Machine Learning\/} 8:229--256.

\bibitem[{Young et~al.(2013)Young, Gasic, Thomson, and Williams}]{young2013}
S.~Young, M.~Gasic, B.~Thomson, and J.~Williams. 2013.
\newblock {POMDP}-based statistical spoken dialog systems: A review.
\newblock {\em Proc. {IEEE}\/} 101(5):1160--1179.

\end{thebibliography}

\appendix
\section{Specifications of restaurant search in DTSC2}


\begin{description} \label{tab:constraints_requests}
    \item[Constraint slots] area, type of food, price range.

    \item[Request slots] area, type of food, address, name, price range, postcode, signature dish, phone number
\end{description}

\begin{table}[h!]
\begin{center}
\caption{Summary actions.}
    \begin{tabu}to\linewidth{@{}X[c,1.5]X[l,5]@{}}
\toprule
\textbf{Action} & \textbf{Description} \\ \midrule \midrule
Cannot help & No restaurant in the database matches the user's constraints. \\ \midrule
Confirm  Domain & Confirm that the user is looking for a restaurant. \\ \midrule
Explicit  Confirm & Ask the user to confirm a piece of information. \\ \midrule
Offer & Propose a restaurant to the user. \\ \midrule
Repeat & Ask the user to repeat. \\ \midrule
Request & Request a slot from the user. \\ \midrule
Select & Ask the user to select a value
 between two propositions
 (\textit{e.g.} select between Italian and Indian). \\ \bottomrule
\end{tabu}
\label{tab:summary_actions}
\end{center}
\end{table}

\begin{table}[h!]
\begin{center}
\caption{User actions.}
    \begin{tabu}to\linewidth{@{}X[2,c]X[5]@{}}
\toprule
\textbf{Action} & \textbf{Description} \\ \midrule \midrule
Deny & Deny a piece of information. \\ \midrule
Null & Say nothing. \\ \midrule
Request More & Request more options. \\ \midrule
Confirm & Ask the system to confirm \\
& a piece of information. \\ \midrule
Acknowledge & Acknowledge. \\ \midrule
Affirm & Say yes. \\ \midrule
Request & Request a slot value. \\ \midrule
Inform & Inform the system of a slot value. \\ \midrule
Thank you & Thank the system. \\ \midrule
Repeat & Ask the system to repeat. \\ \midrule
Request & Request alternative \\
Alternatives & restaurant options.\\ \midrule
Negate & Say no. \\ \midrule
Bye & Say goodbye to the system. \\ \midrule
Hello & Say hello to the system. \\ \midrule
Restart & Ask the system to restart\linebreak the dialogue. \\ \bottomrule
\end{tabu}
\label{tab:user_acts}
\end{center}
\end{table}

\end{document}